\documentclass[runningheads]{llncs}
\usepackage{graphicx}
\usepackage{comment}
\usepackage{amsmath,amssymb} 
\usepackage{color}
\usepackage{subfigure}
\usepackage{setspace}
\usepackage{mathrsfs}
\usepackage{bm}
\usepackage{hyperref}
\usepackage{multirow}

\begin{document}
\pagestyle{headings}
\mainmatter
\def\ECCVSubNumber{3429}  
\newcommand{\etal}{et al. }
\newcommand{\ie}{i.e.}
\newcommand{\eg}{e.g.}
\newcommand{\etc}{etc.}
\title{Multiple Sound Sources Localization from Coarse to Fine} 

\titlerunning{Multiple Sound Sources Localization from Coarse to Fine}
%
\author{Rui Qian\inst{1} \and
Di Hu\inst{2} \and
Heinrich Dinkel\inst{1} \and Mengyue Wu\inst{1} \and\\ Ning Xu\inst{3} \and Weiyao Lin\inst{1}\thanks{∗Corresponding Author, Email: wylin@sjtu.edu.cn}}
\authorrunning{R. Qian et al.}
%
\institute{Shanghai Jiao Tong University, China \and
Baidu Research, China \and
Adobe Research, USA\\
\email{\{qrui9911,wylin,richman,mengyuewu\}@sjtu.edu.cn\\
hudi04@baidu.com\\
nxu@adobe.com}}
\maketitle

\begin{abstract}
How to visually localize multiple sound sources in unconstrained videos is a formidable problem, especially when lack of the pairwise sound-object annotations. To solve this problem, we develop a two-stage audiovisual learning framework that disentangles audio and visual representations of different categories from complex scenes, then performs cross-modal feature alignment in a coarse-to-fine manner. Our model achieves state-of-the-art results on public dataset of localization, as well as considerable performance on multi-source sound localization in complex scenes. We then employ the localization results for sound separation and obtain comparable performance to existing methods. These outcomes demonstrate our model's ability in effectively aligning sounds with specific visual sources. Code is available at \url{https://github.com/shvdiwnkozbw/Multi-Source-Sound-Localization}.
\end{abstract}

\section{Introduction}

Humans usually perceive the world through information in different modalities, \eg, vision and hearing. By leveraging the relevance and complementary between audio and vision, humans can clearly distinguish different sound sources and infer which object is making sound. 
In contrast, machines have been proven capable of separately processing audio and visual information using deep neural networks. 
But can they benefit from joint audiovisual learning?

Works in recent years mainly focus on establishing multi-modal relationship based on temporally synchronized audio and visual signals \cite{L3,soundnet,soundsupv,cotrain}.
This synchronization in video-level becomes the correspondence that is whether audio and visual signals originate from the same video, which works effectively for simple scenes \cite{objsound,avscene}, i.e., the single-source conditions. 
However, in unconstrained videos, various sounds are usaully mixed, where the video-level supervision is too coarse to provide the precise alignment between each sound and visual source pair. 
To tackle this problem, \cite{clustering,curriculum} establish audiovisual clusters to associate sound-object pairs, but require to pre-determine the number of clusters, which becomes difficult in an unconstrained scenario, thus greatly affects alignment performance.

\begin{figure}[t]
    \centering
    \includegraphics[width=0.9\linewidth]{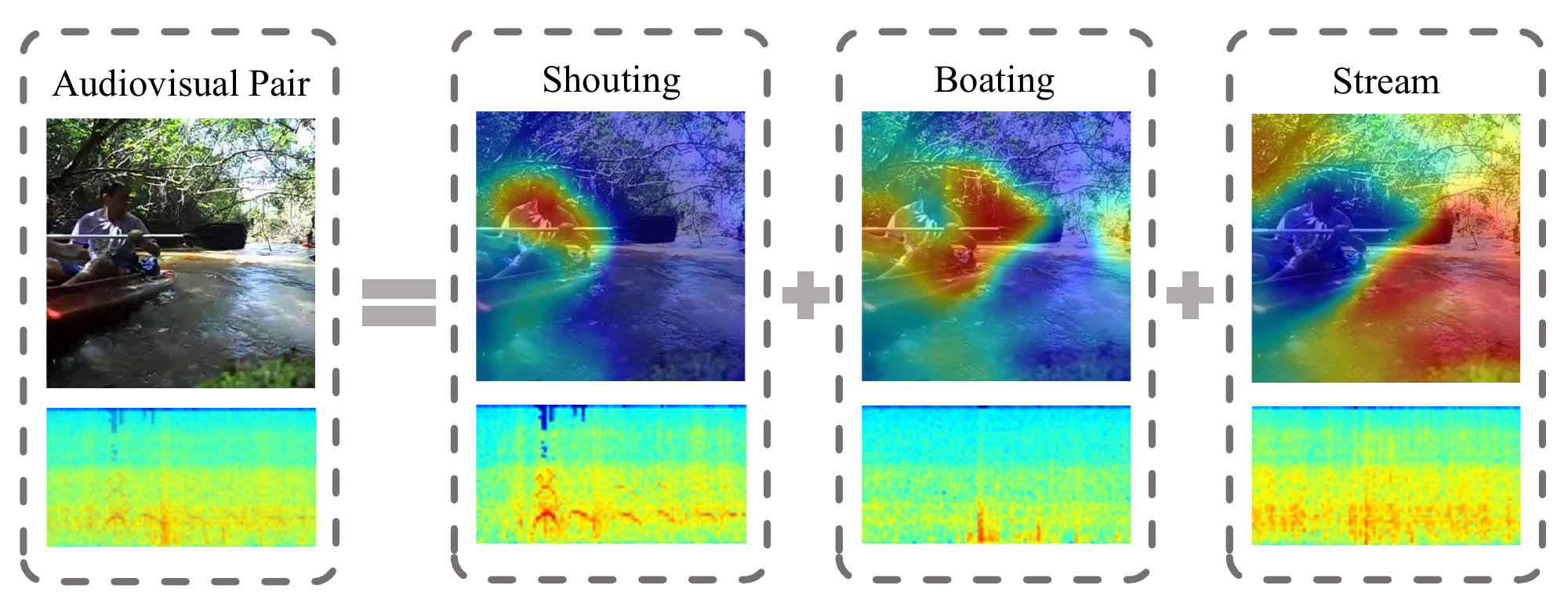}
    \caption{Our model separates a complex audiovisual scene into several simple scenes. The figure shows the input audiovisual pair majorly consists of three elements: a man shouting, sound of boating from the boat and paddle, sound of a water stream. This disentanglement simplifies a complex scenario and generates several one-to-one audiovisual associations.}
    \label{fig: sep}
\end{figure}

Some works further apply audiovisual learning into a series of downstream tasks (\eg, sound localization, sound separation) and exhibit promising performance \cite{attention,avscene,curriculum,sop,co-separation,rouditchenko2019self,zhou2020sep}. 
Regarding previous works on sound localization, \cite{objsound,avscene,attention} mainly focus on simple scenes, usually unable to find source-specific objects from mixed audio, while \cite{vehicle,2.5d,gan2019look} employ stereo audio as prior, which contains location information but is difficult to obtain.
Additionally, existing evaluation pipelines also lack the ability to measure sound localization performance in multi-source scenarios. 
For sound separation, \cite{sop} uses the entire coarse visual scene as guidance, while \cite{som,co-separation,gan2020music} rely on extra motion or detection results to improve performance.

To sum up, existing dominant methods mostly lack the ability to analyze complex audiovisual scenes, and fail to effectively utilize the latent alignment between sound and visual source pairs in unconstrained videos. This is because there are majorly two challenges in complex audiovisual scene analysis: one is how to distinguish different sound-sources, the other is how to ensure the established sound-object alignment is fairly satisfactory without one-to-one annotations.
To address these challenges, we develop a two-stage audiovisual learning framework. At the first stage, we employ a multi-task framework consisting of classification and audiovisual correspondence to provide the reference of audiovisual content for the second stage.  
At the second stage, based on the classification predictions, we use the operation of \emph{Class Activation Mapping} (CAM) \cite{cam,grad-cam,grad-cam++} to extract class-specific feature representations as the potential sound-object pairs (Fig.~\ref{fig: sep}), then perform alignment in a coarse-to-fine manner, where the coarse correspondence based on category is evolved into the fine-grained matching in both video- and category-level.

Our main contributions can be summarized as follows: (1) We develop a two-stage audiovisual learning framework. At the first stage, we employ multi-task framework for classification and correspondence learning. At the second stage, we employ the CAM technique to disentangle the elements of different categories from complex scenes for alignment. (2) We propose to establish audiovisual alignment in a coarse-to-fine manner. The coarse-grained step ensures correctness of correspondence in category level, while the fine-grained one establishes video- and category-based sound-object association.
(3) We achieve state-of-the-art results on public sound localization dataset. In the multi-source conditions, according to our proposed class-specific localization metric, our method shows considerable performance compared with several baselines. Besides, the object representation obtained from localization provides valuable visual reference for sound separation.

\section{Related Work}

\subsubsection{Audiovisual Correspondence.}

Although most audiovisual datasets consist of unlabelled videos, the natural correspondence between sound and vision provides essential supervision for audiovisual learning \cite{L3,objsound,avscene,soundnet,soundsupv}. \cite{soundnet,soundsupv} introduced a method to learn feature representation of one modality with supervision from the other in a teacher-student manner.
Arandjelovic and Zisserman \cite{L3} viewed audiovisual correspondence (AVC) as the supervision for audiovisual representation learning.
\cite{avscene} adopted temporal synchronization as self-supervision signal to correlate audiovisual content.
But these methods mostly fail to process complex scene with multiple sound sources.
Hu \etal \cite{clustering,curriculum} used clustering to associate latent sound-object pairs, but its performance greatly relies on predefined number of clusterings.
Our multi-task framework simultaneously treats unimodal content label and audiovisual correspondence as supervision, then performs class-specific audiovisual alignment under complex scenes.

\subsubsection{Sound Localization in Visual Scenes.}

Recent methods for localizing sound in visual context mainly focus on joint modeling of audio and visual modalities \cite{objsound,avscene,attention,ave,clustering,som,sop}.
In \cite{objsound,avscene}, authors performed sound localization through audiovisual correspondences. 
\cite{attention} proposed an attention mechanism to capture primary areas in a semi-supervised or unsupervised setting. Tian \etal\cite{ave} leveraged audio-guided visual attention and temporal alignment to find semantic regions corresponding to sound sources. Hu \etal\cite{clustering,curriculum} established audiovisual clustering to localize sound makers. Zhao \etal\cite{sop,som} employed a self-supervised framework to simultaneously achieve sound separation and visual grounding. Although \cite{sop,som} can separate sound given visual sound source, they require single-source samples to achieve mix-and-separate training. In contrast, our model is directly trained on unconstrained videos, and can precisely localize visual source of different sounds in complex scenes.

\subsubsection{CAM for Weakly-Supervised Localization.}

CAM was proposed by Zhou \etal\cite{cam} to localize objects with only holistic image labels.
This approach employs a weighted sum of the global average pooled features at the last convolutional layer to generate class-specific saliency maps, but can only be applied to fully-convolutional networks due to modification of network architectures.
To generalize CAM and improve visual explanations for convolutional networks, Grad-CAM \cite{grad-cam} and Grad-CAM++ \cite{grad-cam++} were proposed. These two gradient-based methods can achieve weakly-supervised localization with arbitrary off-the-shelf CNN architectures and require no re-training.

Some previous works on audiovisual learning have adopted CAM or similar methods to localize sound producers \cite{sop,objsound,avscene}. Arandjelovic \etal \cite{objsound} performed max pooling on predicted score map over all spatial grids, and used obtained correspondence score for training on AVC task. Owens \etal \cite{avscene} adopted audiovisual synchronization as training supervision, and employed CAM to measure the likelihood of a patch to be sound source.
However, they only use CAM at the final step to measure the relationship between two modalities. Our method employs CAM to disentangle audio and visual features of different sounding objects, achieving fine-grained audiovisual alignment.

\section{Approach}

\begin{figure}[t]
    \centering
    \includegraphics[width=0.9\linewidth]{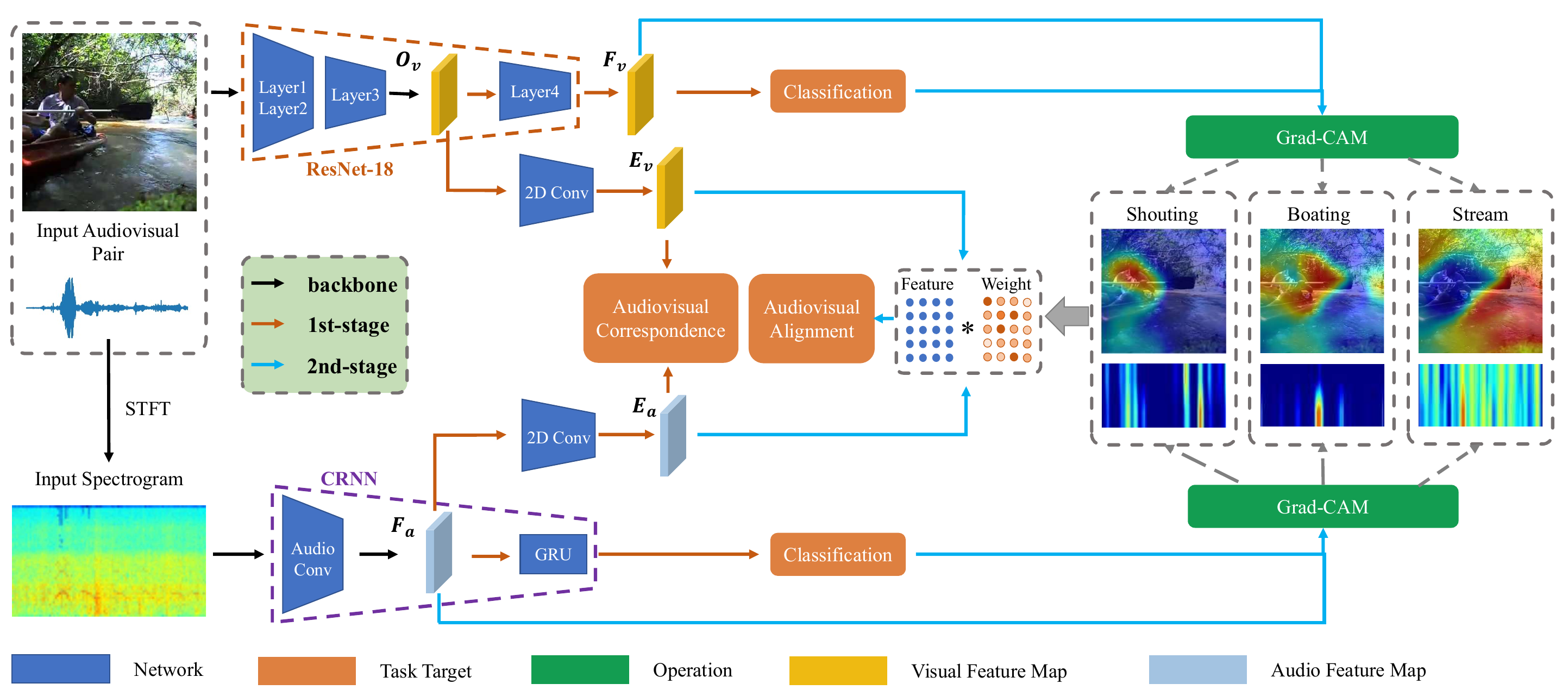}
    \caption{An overview of our two-stage audiovisual learning framework. At the first stage, our model extracts deep features from the audio and visual streams, then performs classification and video-level correspondence. At the second stage, our model disentangles representations of different classes and implements a fine-grained audiovisual alignment.}
    \label{fig: flow}
\end{figure}

Our two-stage framework is illustrated in Fig.~\ref{fig: flow}. At the first stage, we employ multi-task learning for classification and video-level audiovisual correspondence. At the second stage, the audiovisual feature maps and classification predictions are fed into Grad-CAM \cite{grad-cam} module to disentangle class-specific features on both modalities, based on which we employ valid representations to perform fine-grained audiovisual alignment.
\subsection{Multi-Task Training Framework}

Given audio and visual (image) messages $\{a_i,v_i\}$ from \textit{i-th} video, we can obtain the category labels from annotated video tags or predictions of pretrained models, as well as the natural audiovisual correspondence.
To leverage these two types of supervision, we employ a multi-task learning model.
This model consists of audio and visual learning backbones, classification network and an audiovisual correspondence network, as shown in Fig.~\ref{fig: flow}.
Specifically, we adopt CRNN \cite{crnn}, composed of 2D convolutions and a GRU, to process audio spectrograms, and use ResNet-18 \cite{resnet} to extract deep features from video frames.

\subsubsection{Classification on Two Modalities.}

To perform classification with audio and visual messages $\{a_i,v_i\}$, we adopt video tags or predicted pseudo labels from pretrained models as supervision. 
Considering the sound-object alignment to be established, we employ the same categories for both modalities. We denote $C$ as the number of class and $c$ as the \textit{c-th} class. 
Considering there are multiple sound sources contained in the video, multi-label binary cross entropy loss is considered for classification:
\begin{align}
    \label{con: cls}
    L_{cls} = \mathcal{H}_{bce}(\bm{y}_{a_i},\bm{p}_{a_i}) + \mathcal{H}_{bce}(\bm{y}_{v_i},\bm{p}_{v_i}),
\end{align}
where $\mathcal{H}_{bce}$ is the binary cross-entropy loss for multi-label classification, $\bm{y}$ and $\bm{p}$ are the annotated class labels and corresponding predicted probability respectively, $\bm{y}\in \{0,1\}^C$, $\bm{p} \in \left[0,1\right]^C$.

\begin{figure}
    \centering
    \includegraphics[width=0.9\linewidth]{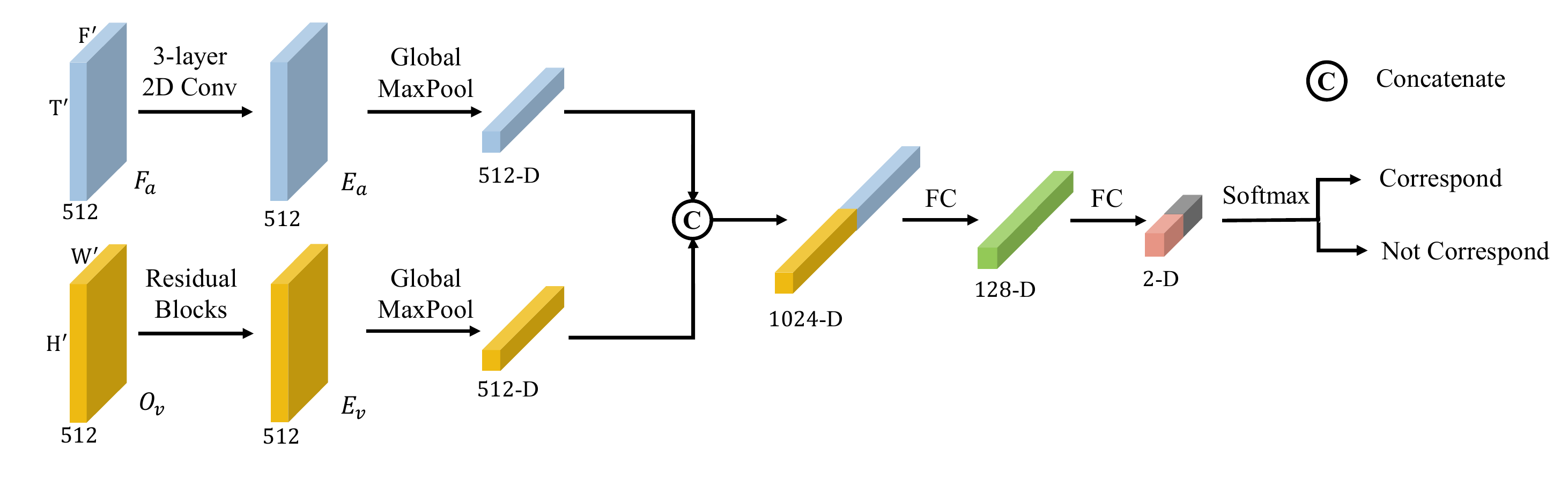}
    \caption{Details for audiovisual correspondence learning network. For audio stream, the 3-layer 2D convolutions are listed as: (1) 3$\times$1$\times512$, with dilation 2 on time dimension, (2) 1$\times$2$\times$512, with stride 2 on frequency dimension, (3) 3$\times$1$\times$512, each followed with a batch normalization layer and ReLU activation. For visual stream. the layer settings for residual blocks are the same as layer4 in ResNet-18, but the weights are not shared.}
    \label{fig: coal}
\end{figure}

\subsubsection{Audiovisual Correspondence Learning.}
\label{sec: avc}
Similar to \cite{L3}, audiovisual correspondence learning is viewed as a two-class classification problem, i.e., corresponding or not. And the network shown in Fig.~\ref{fig: coal} is employed for achieving this learning task.
Specifically, we take audio features before GRU in CRNN and visual outputs from layer3 of ResNet-18 as inputs\footnote{We choose $F_a$ and $O_v$ for two reasons: we can obtain more fine-grained local features and achieve easier training process.}, \ie, $F_a$ and $O_v$ in Fig.~\ref{fig: flow}. 
Through a series of convolution and pooling operation in Fig.~\ref{fig: coal}, we can get 512-D audio and visual features. Then, these two 512-D features are concatenated into one 1024-D vector and passed through two fully-connected layers of 1024-128-2. The 2-D output with softmax regression aims to determine whether audio and vision correspond.

$\{a_i,v_i\}$ from \textit{i-th} video are viewed as corresponding pair, then we random select a different video $j$ and use its image $v_j$ to construct mis-corresponding pair $\{a_i,v_j\}$. 
The learning objective can be written as:
\begin{align}
    \label{con: cor}
    L_{avc} = \mathcal{H}_{cce}(\bm{\delta},\bm{q}),
\end{align}
where $\mathcal{H}_{cce}$ is the categorical cross entropy loss, $\bm{q}\in \left[0,1\right]^2$ is the predicted output, $\bm{\delta}$ is the class indicator, $\bm{\delta}=\left(0,1\right)$ for correspondence while $\bm{\delta}=\left(1,0\right)$ for not.
For multi-task learning, we take $L_{mul}$ as final loss function, $\lambda$ is the hyperparameter of weighting:
\begin{align}
    \label{con: mul}
    L_{mul} = L_{cls} + \lambda L_{avc}.
\end{align}
After training with multi-task objective, we could achieve coarse-grained audiovisual correspondence in the category level.

\subsection{Audiovisual Feature Alignment}

In this section, we propose to disentangle feature representations of different categories based on the classification predictions and implement fine-grained audiovisual alignment with the video- and category-based sound-object association.

\subsubsection{Disentangle Features by Grad-CAM.}
\label{sec: grad-cam}

Inspired by \cite{cam,grad-cam,grad-cam++}, CAM method can generate class-specific localization maps, which measures the importance of each spatial grid on the feature map to specific categories, through classification task. Hence, it is feasible for us to disentangle feature representations of different classes based on the predictions in \ref{sec: avc}.

Specifically, we leverage the operation of Grad-CAM \cite{grad-cam} to perform disentanglement. For simplicity, we use $r\in \{a,v\}$ to represent audio or visual modality. Given the feature map activations of the last convolutional layer, $F_r$, and the output of classification branch without activation for class $c$, $\hat{p_r^c}$, we calculate the class-specific map $W_r^c$, \ie,
\begin{align}
    W_r^c = \text{Grad-CAM}(F_r,\hat{p_r^c}).
    \label{con: grad-cam}
\end{align}
Then we take class-specific map $W_r^c$, \ie, the visualized heatmap in Fig.~\ref{fig: flow}, as weights to perform weighted global pooling over the feature map $E_r(u,v)$ to obtain class-aware representation\footnote{We find that directly using $F_r$ with the weights $W_r^c$ is difficult to perform alignment objective, but by performing weighted pooling on $E_r$, we achieve easier training and faster convergence.}, where $u$ and $v$ are the map entries. That is:
\begin{align}
    f_r^c = \frac{\sum_{u,v} E_{r}(u,v)W^c_{r}(u,v)}{\sum_{u,v} W^c_{r}(u,v)}.
    \label{con: pooling}
\end{align}

Finally, we get $C$ 512-D vectors as the feature representation of all the categories. And $\{f_{a_i}^m|m=1,2,...,C\}$ and $\{f_{v_i}^n|n=1,2,...,C\}$ are as the set of audio and visual class-specific feature representations for \textit{i-th} video. We use them for fine-grained feature alignment in next step.

\subsubsection{Fine-Grained Audiovisual Alignment.}


To effectively establish audiovisual alignment with disentangled features, there are potentially two ways. One is to treat all audio and visual features of the same class in a batch as positive pairs for alignment, the other is to only take pairs of the same class from the same video as positive. As each category contains various entities (\eg, the human category contains audio and visual patterns of baby, sportsman, old man \etc), in order to reduce the interference among different entities, we choose the latter one to 
acquire the positive pairs with higher quality.


To effectively compare the class-specific audio and visual representation, \ie, $f_{a_i}^m$ and  $f_{v_j}^n$, we project them into a shared embedding space via two fully-connected layers of 512-1024-128, respectively. Then we compare the projected features with Euclidean distance, 
\begin{align}
    D(f_{a_i}^m, f_{v_j}^n) = ||g_a(f_{a_i}^m)-g_v(f_{v_j}^n)||_2,
    \label{con: dist}
\end{align}
where $g_a$ and $g_v$ are the fully-connected layers for audio and visual modalities, respectively. We then adopt contrastive loss \cite{contrast} to implement sound-object alignment. 
The loss function is written as\footnote{In practice, a threshold over all the class predictions is considered to select valid categories.}
\begin{align}
\begin{split}
    L_{ava} = \sum_{i,j=1}^N\sum_{m}\sum_{n} (\delta_{i=j}^{m=n} D^2(f_{a_i}^m, f_{v_j}^n)+\\
    (1-\delta_{i=j}^{m=n})max(\Delta-D(f_{a_i}^m, f_{v_j}^n), 0)^2),
    \label{con: contrast}
\end{split}
\end{align}
where $\delta_{i=j}^{m=n}$ indicates whether the audiovisual pair is positive, \ie, $\delta_{i=j}^{m=n}=1$ when $i=j$ and $m=n$, otherwise 0. $\Delta$ is a margin hyper-parameter.

\subsection{Sound Localization and Its Application in Separation}

In this section, we use our method to visually localize sounds, and adopt localization results as object representation to guide sound separation.

\subsubsection{Visual Localization of Sounds.}
\label{sec: localization}
In this task, we aim to visually localize sounds by generating source-aware localization maps.
To leverage the established alignment to associate sounds with objects, 
the visual feature map $E_{v_i}$ of testing image is firstly projected into the shared embedding space via $g_v$ in Eq.~\ref{con: dist}, then compared with the disentangled \textit{c-th} class audio features $f_{a_i}^c$ through Eq. \ref{con: dismap},
\begin{align}
    K_i^c(u,v) = -||g_a(f_{a_i}^c)-g_v(E_{v_i})(u,v)||_2.
    \label{con: dismap}
\end{align}
Note that $g_v$ in Eq. \ref{con: dismap} is transformed into $1\times 1$ convolutions with parameters unchanged.
The obtained $K_i^c\in \mathbb{R}^{U\times V}$ reveals how likely a specific region in the visual scene $v_i$ is the \textit{c-th} visual source of sound $a_i$. Then, $K_i^c$ is normalized and resized to the original image size to be the final localization maps for sound source in the \textit{c-th} class. Further, the localization results with class label can be used to evaluate sound localization performance in multi-source conditions.

\subsubsection{Sound Source Separation.}
\label{sec: sep}
To evaluate the effectiveness of our sound localization results, we use localized objects to guide sound separation.
To generate the visual source guidance for the sound belonging to \textit{c-th} class, we perform weighted global pooling over the feature map $E_{v_i}$ w.r.t. the localized visual source $K_i^c$, similar to Eq. \ref{con: pooling}. Then, following \cite{sop}, we adopt the same mix-and-separate learning framework, and take U-Net \cite{unet} to process mixed audio spectrogram, where the visual representation of object in \cite{sop} is replaced by our automatically determined visual source guidance. Finally, the output of masked spectrogram w.r.t. the visual source is converted into audio waveform via inverse short-time Fourier transform. More details about the processing can be found in \cite{sop}. 

\section{Experiments}

\subsection{Datasets}

\subsubsection{SoundNet-Flickr.}

This dataset was proposed in \cite{soundnet}, containing over 2 million unconstrained videos from Flickr. Following \cite{L3,attention}, we adopt one 5-second audio clip and its corresponding image as an audiovisual pair, and no extra supervision is used for training. For quantitative evaluation of sound localization, the human-annotated subset of SoundNet-Flickr \cite{attention} is adopted. In our setting, a random subset of 10k pairs is used for training, and 250 annotated pairs for testing.

\subsubsection{AudioSet.}

AudioSet consists of mainly 10-second video clips, many containing multiple sound sources, divided into 632 event categories.
Following \cite{MIML,co-separation}, we only consider sounds from 15 musical instruments  extracted from the ``unbalanced" split for training and from the ``balanced" split for testing. 
Since this subset provides musical scenes with multiple sound sources, some of poor quality, it is proper and also challenging for multi-source sound localization evaluation.
We extract video frames at 1 fps, and employ the well-trained Faster RCNN detector w.r.t. these 15 instruments \cite{co-separation} to provide object locations (bounding boxes), which is then used as the evaluation reference for the sound localization. Finally, we get 96,414 10-second clips for training, and 4503 ones for testing\footnote{Since AudioSet only provides clip-level audio labels, we can only ensure that labelled sounds appear in the clip. Thus we adopt the whole 10-second audio clip with one randomly selected frame from video as a pair.}.

\subsubsection{MUSIC.}

MUSIC dataset consists of 685 untrimmed videos, with 536 musical solo and 149 duet, containing 11 categories of musical instrument. Since this dataset contains less noise and cleaner than AudioSet, it is more proper to train sound separation models. Following \cite{sop}, we set the first/second video of each category as validation/test set, and use the rest for training. But some videos have been removed on YouTube, we finally get 474 solo and 105 duet videos in total.

\subsection{Implementation Details}
\label{sec: detail}
Our audiovisual learning model is implemented in PyTorch. 
We pretrain CRNN \cite{crnn} and ResNet-18 \cite{resnet} model as audio and visual feature extractors.
The CRNN is pretrained on a subset of the unbalanced AudioSet corpus, encompassing 700k audio-clips out of the available 2 Million. The ResNet-18 is pretrained on ImageNet.

For all experiments, if not specially mentioned, we sample the audio at 22.05kHz and convert it to log-mel spectrogram (LMS) \cite{melspec}, obtaining 64 frequency bins from a window of 40ms every 20ms using the librosa framework.
Regarding visual input, we resize the image to $256\times256\times3$. Our model is optimized in a two-stage manner.
First, we set $\lambda$ to 1 and train the multi-task model w.r.t. Eq. \ref{con: mul} in section \ref{sec: avc}. Then, we jointly optimize the entire network w.r.t. Eq. \ref{con: mul} and Eq. \ref{con: contrast}. The model is trained by SGD optimizer with momentum 0.9 and starting learning rate $1\times10^{-3}$. We set learning rate for two backbones to $1\times10^{-4}$. The learning rate is decreased by 0.1 every 20 epochs.

\subsection{Sound Localization}

\subsubsection{Sound Localization on SoundNet-Flickr.}

\begin{figure}[t]
    \subfigure[speech with gunfire]{
    \includegraphics[width=0.105\linewidth]{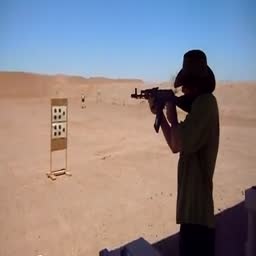}\hspace{-1mm}
    \includegraphics[width=0.105\linewidth]{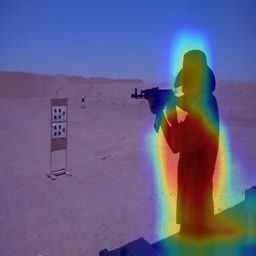}\hspace{-1mm}
    \includegraphics[width=0.105\linewidth]{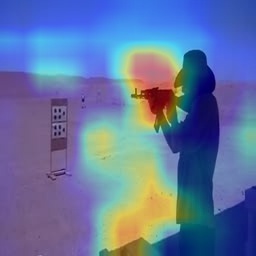}
    \label{fig: human-gun}}
    \hspace{-4mm}
    \subfigure[cheering with engine]{
    \includegraphics[width=0.105\linewidth]{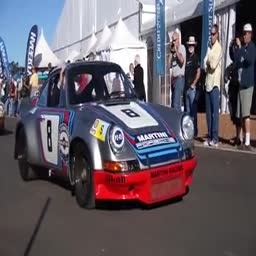}\hspace{-1mm}
    \includegraphics[width=0.105\linewidth]{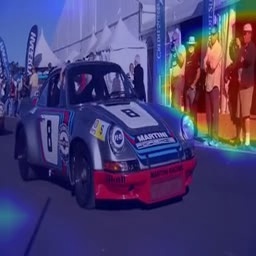}\hspace{-1mm}
    \includegraphics[width=0.105\linewidth]{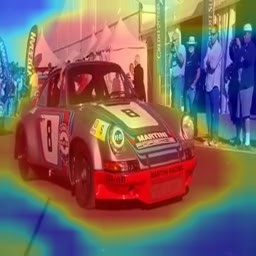}
    }
    \hspace{-4mm}
    \subfigure[shouting with water]{
    \includegraphics[width=0.105\linewidth]{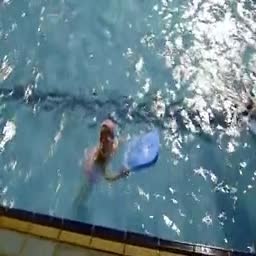}\hspace{-1mm}
    \includegraphics[width=0.105\linewidth]{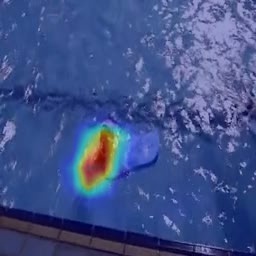}\hspace{-1mm}
    \includegraphics[width=0.105\linewidth]{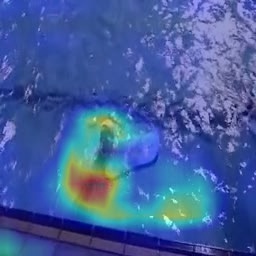}
    }
    \vspace{-3mm}\\
    \subfigure[sports with stadium]{
    \includegraphics[width=0.105\linewidth]{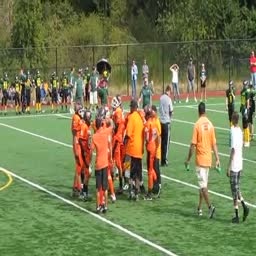}\hspace{-1mm}
    \includegraphics[width=0.105\linewidth]{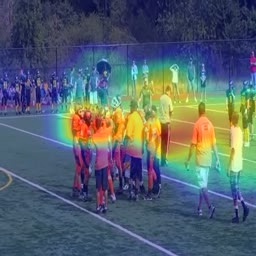}\hspace{-1mm}
    \includegraphics[width=0.105\linewidth]{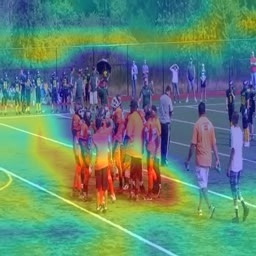}
    }
    \hspace{-4mm}
    \subfigure[speech with motorcycle]{
    \includegraphics[width=0.105\linewidth]{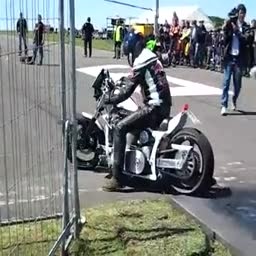}\hspace{-1mm}
    \includegraphics[width=0.105\linewidth]{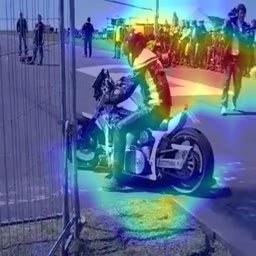}\hspace{-1mm}
    \includegraphics[width=0.105\linewidth]{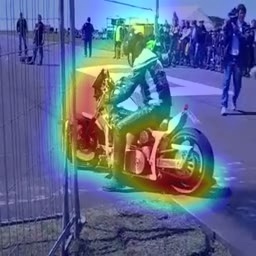}
    }
    \hspace{-4mm}
    \subfigure[yelling with impact]{
    \includegraphics[width=0.105\linewidth]{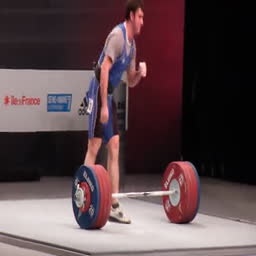}\hspace{-1mm}
    \includegraphics[width=0.105\linewidth]{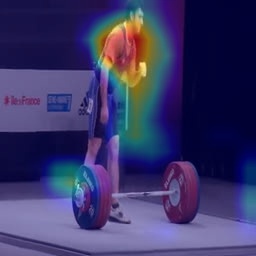}\hspace{-1mm}
    \includegraphics[width=0.105\linewidth]{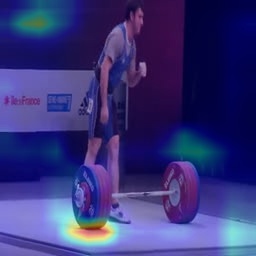}
    \label{fig: human-impact}}
    \caption{We visualize the localization maps corresponding to different elements contained in the mixed sounds of two sources. The results qualitatively demonstrate our model's performance in multi-source sound localization.}
    \label{fig:flickr}
\end{figure}

\begin{figure}
    \centering
    \subfigure[playing violin]{
    \includegraphics[width=0.155\linewidth]{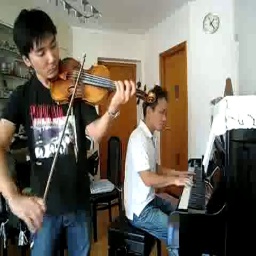}\hspace{-1mm}
    \includegraphics[width=0.155\linewidth]{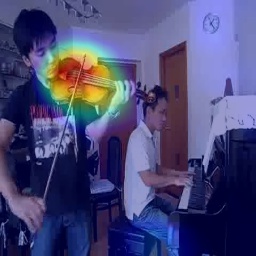}\hspace{-1mm}
    \includegraphics[width=0.155\linewidth]{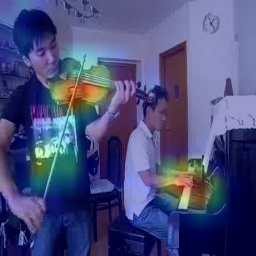}
    \label{fig: sew}}
    \hspace{-2mm}
    \subfigure[yelling sound]{
    \includegraphics[width=0.155\linewidth]{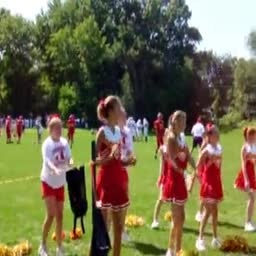}\hspace{-1mm}
    \includegraphics[width=0.155\linewidth]{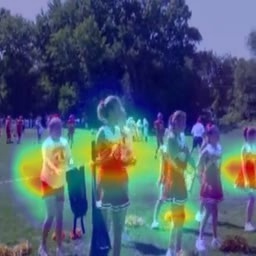}\hspace{-1mm}
    \includegraphics[width=0.155\linewidth]{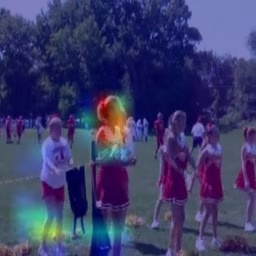}
    \label{fig:cam}}
    \caption{We compare violin and human yelling sound localization results of our model and CAM output of corresponding category, images in each subfigure are listed as: original image, localization result of our method, and result of CAM.}
    \label{fig:3-source}
\end{figure}

In this section, we adopt audiovisual pairs from SoundNet-Flickr \cite{soundnet} for training and evaluation. The videos in this dataset are completely unconstrained and noisy, thus very challenging to localize sound sources. As there are no video tags available, we adopt the first-level labels in AudioSet \cite{audioset} of 7 categories (human sounds, music, animal, sounds of things, natural sounds, source-ambiguous sounds, and environment) as final classification target. We correlate ImageNet labels with these 7 categories by using the similarity of word embeddings \cite{glove} and conditional probabilities between labels of these two datasets, more details on this are in the supplemental materials. The pseudo labels are generated based on the prediction of pretrained CRNN and ResNet-18 model.
For evaluation, we disentangle class-specific features on audio stream and localize corresponding sound source on each spatial grid of visual feature maps.

To effectively present our model's ability of category-level disentanglement and fine-grained alignment, we visualize video frames with localization maps in Fig.~\ref{fig:flickr}. Unlike \cite{attention} which inputs different types of audio to demonstrate interactive sound localization, we input a mixed audio containing multiple sources to generate class-specific localization responses. For example, in Fig.~\ref{fig: human-gun}, when input audio clip contains human speaking and sound of gunfire, our model automatically separates these two parts and respectively highlights the person and gun area. Besides sounds with clear visual sources, for source-ambiguous sound like impact, our model accurately captures the contact surface as shown in Fig.~\ref{fig: human-impact}. More examples are shown in the supplemental material.

The comparison between our model and CAM is shown in Fig.~\ref{fig:3-source}. First, our method can generally associate sounds with specific sources. In Fig. \ref{fig: sew}, the violin is making sound while the piano is silent, and our method accurately distinguishes these two objects that belong to the same category of ``music", which surpasses the category-based localization technique of CAM.
Second, compared with CAM, Fig.~\ref{fig:cam} shows that our model can precisely localize the position of human by listening to the yelling sound but CAM somewhat fails to achieve this only with human category information. More comparison examples are shown in the supplemental material.

Further we implement quantitative evaluation on 249 pairs from human annotated subset of SoundNet-Flickr \cite{attention}. Consensus Intersection over Union (cIoU) and Area Under Curve (AUC) \cite{attention} are employed as evaluation metrics. To evaluate the localization response to the entire audio, we perform weighted summation over valid categories as final localization map, where the weights are the normalized predicted probabilities. Table~\ref{tab: flickr} shows the results for different methods, all of which are trained in an unsupervised manner. Despite that most audiovisual pairs in test set are of single-source, our model still outperforms Attention \cite{attention} and DMC \cite{clustering} by a large margin, and is slightly better than CAVL \cite{curriculum}. but note that CAVL is trained on single-source videos while our model is trained on unconstrained ones, which poses greater challenge in the joint audiovisual learning. This result demonstrates that our fine-grained alignment effectively facilitates audiovisual learning with unconstrained videos. Due to limited computing resources, we did not try very large training data size like 144K as \cite{attention}, but the result on 20K training data has shown the performance is increased with the number of training data.
\setlength{\tabcolsep}{4mm}{
\begin{table}[]
    \centering
    \caption{Quantitative localization results on SoundNet-Flickr subset, cIoU and AUC are reported (results of other methods are directly reported from \cite{curriculum}).}
    \begin{spacing}{1.00}
    \begin{tabular}{c | c c}
    \hline
    Methods & cIoU@0.5 & AUC  \\
    \hline
    Random & 7.2 & 30.7 \\
    Attention 10K\cite{attention} & 43.6 & 44.9 \\
    DMC AudioSet\cite{clustering} & 41.6 & 45.2 \\
    CAVL AudioSet\cite{curriculum} & 50.0 & 49.2 \\
    Ours 10K & \textbf{52.2} & \textbf{49.6} \\
    \hline
    Ours 20K & 53.8 & 50.6 \\
    \hline
    \end{tabular}
    \end{spacing}
    \label{tab: flickr}
\end{table}}

\subsubsection{Multi-Source Localization on AudioSet.}
\label{sec: eval}
Since existing methods of sound localization evaluation are mainly for single-source scenes, We propose a quantitative evaluation pipeline for multi-source sound localization in complex scenes.
We adopt a subset of AudioSet covering 15 musical instruments for training and testing. 

To evaluate the model's ability of separating sounds of different instruments and aligning them with corresponding visual sources, we use cIoU and AUC metric in a class-aware manner. Different from class-agnostic score map used in \cite{attention}, our method uses the detected bounding boxes of Faster RCNN to indicate the localization of sounding objects\footnote{We have filtered out those silent detected objects.}, each box is labelled as one specific category of music instruments, \ie, $C=15$ on this dataset. Next, we calculate cIoU scores (\eg, with threshold 0.5) on each valid sound source and take an average. Final cIoU\_class on each frame can be calculated by
\begin{align}
    {\rm cIoU\_class} = \frac{\sum_{c=1}^{C}\theta_c cIoU_c}{\sum_{c=1}^{C}\theta_c},
\end{align}
where $c$ indicates the class index of instruments, $\theta_c=1$ if instrument of class $c$ makes sounds, otherwise 0. In this way, only when the model is able to establish class-specific association between sounds and objects, the evaluation score of cIoU\_class will become high.
\setlength{\tabcolsep}{4pt}
\begin{table}
    \centering
    \caption{Quantitative localization results on AudioSet of different difficulty levels. The cIoU\_class threshold is 0.5 for level-1 and level-2, but 0.3 for level-3. Note that $\dagger$AVC method is evaluated in a class-agnostic way.}
    \begin{spacing}{1.00}
    \begin{tabular}{c|c c|c c|c c}
    \hline
    \multirow{2}*{Methods} & \multicolumn{2}{|c|}{level-1} & \multicolumn{2}{|c|}{level-2} & \multicolumn{2}{|c}{level-3} \\
    \cline{2-7}
    ~ & cIoU\_class & AUC & cIoU\_class & AUC & cIoU\_class@0.3 & AUC \\
    \hline
    $\dagger$AVC & 24.8 & 32.0 & 4.27 & 23.6 & 5.3 & 14.9 \\
    Multi-task & 20.6 & 29.5 & 2.37 & 17.4 & 10.5 & 17.8 \\
    Ours & \textbf{32.8} & \textbf{38.3} & \textbf{6.16} & \textbf{23.9} & \textbf{21.1} & \textbf{22.0} \\ \hline
    \end{tabular}
    \end{spacing}
    \label{tab: audioset}
\end{table}

To clearly present the effectiveness of our audiovisual alignment, we further divide the testing set into different difficulty levels based on the number of categories of sounding instruments, which results in 4,273 pairs of single-source (level-1), 211 pairs of two-source (level-2) and 19 pairs of three-source (level-3). 
As our model is a two-stage learning method, consisting of multi-task learning and fine-grained alignment, to validate the contribution of each of them, we conduct an ablation study with two baselines. The two baselines are (1) AVC: only using video-level audiovisual correspondence for training and inferring the sound locations in a class-agnostic way. (2) multi-task learning: using both of classification and audiovisual correspondence for training and inferring the sound locations with the coarse-grained audiovisual correspondence.
Table~\ref{tab: audioset} shows the localization results on different difficulty levels. Note that, as AVC method is not provided with any category information, we evaluate it in a class-agnostic way.
From the results, we have several observations. First, using AVC to localize sound in a class-agnostic way is effective with limited sound sources, but fails when more objects make sounds. This is because the video-level correspondence is too coarse to provide sound-object association in complex scenes. Second, although AVC takes a much looser evaluation metric of class-agnostic, it is still worse than the multi-task method on level-3, which reveals introduced classification helps to distinguish sounds of different sources. 
Third, our method with audiovisual alignment significantly outperforms two baselines and is robust on all difficulty levels. It demonstrates that our feature disentanglement and fine-grained alignment is effective to establish one-to-one association in both single-source and multi-source scenes.

\begin{figure}
    \centering
    \subfigure[]{
    \includegraphics[width=0.135\linewidth]{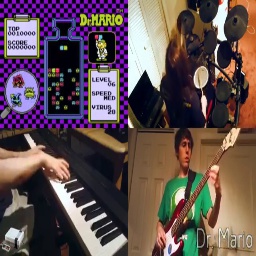}
    }\hspace{-2mm}
    \subfigure[Guitar]{
    \includegraphics[width=0.135\linewidth]{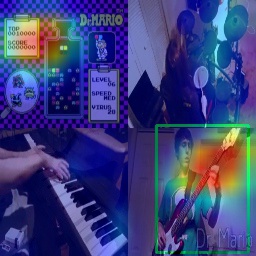}\hspace{-1mm}
    \includegraphics[width=0.135\linewidth]{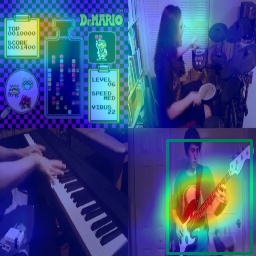}\hspace{-1mm}
    \includegraphics[width=0.135\linewidth]{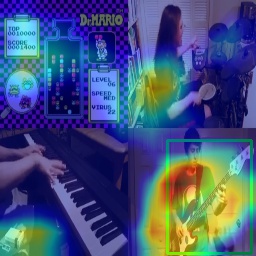}
    }\hspace{-2mm}
    \subfigure[Piano]{
    \includegraphics[width=0.135\linewidth]{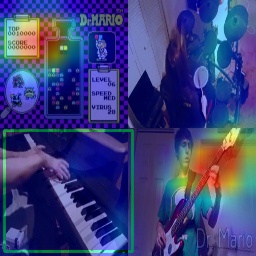}\hspace{-1mm}
    \includegraphics[width=0.135\linewidth]{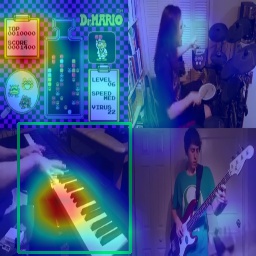}\hspace{-1mm}
    \includegraphics[width=0.135\linewidth]{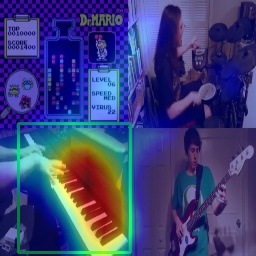}
    }\\
    \vspace{-3mm}
    \subfigure[]{
    \includegraphics[width=0.135\linewidth]{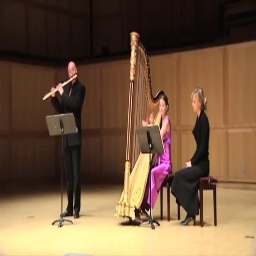}
    }\hspace{-2mm}
    \subfigure[Harp]{
    \includegraphics[width=0.135\linewidth]{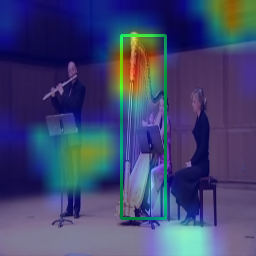}\hspace{-1mm}
    \includegraphics[width=0.135\linewidth]{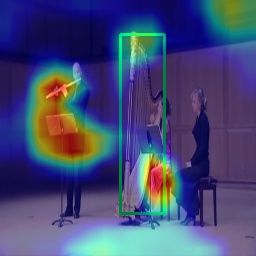}\hspace{-1mm}
    \includegraphics[width=0.135\linewidth]{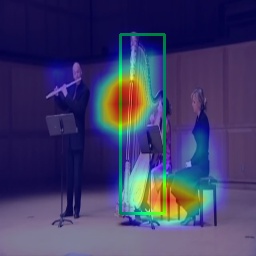}
    }\hspace{-2mm}
    \subfigure[Flute]{
    \includegraphics[width=0.135\linewidth]{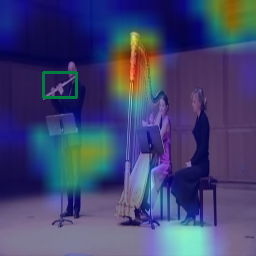}\hspace{-1mm}
    \includegraphics[width=0.135\linewidth]{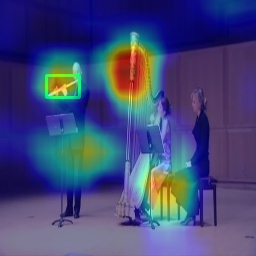}\hspace{-1mm}
    \includegraphics[width=0.135\linewidth]{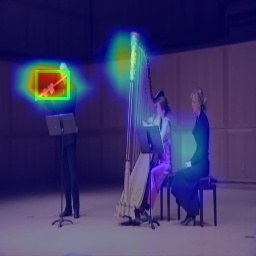}
    \label{fig: flute}}
    \caption{We visualize some examples in AudioSet level-2. The localization maps in each subfigure are listed from left to right: AVC, Multi-task, Ours. The green boxes are detection results of Faster RCNN.}
    \label{fig: audioset}
\end{figure}

We visualize some localization maps for the scenes in level-2 w.r.t. three different methods: AVC, Multi-task and Ours in Fig.~\ref{fig: audioset}. It is clear that our method can generally associate sounds with specific instruments. For example, our method precisely focuses on the tiny area where the flute locates, while the other two associate flute sound with visual object of harp.

\subsection{Sound Separation}

In this task, we use localized objects as visual guidance to perform sound source separation, and evaluate it on MUSIC dataset. Following \cite{sop}, we sub-sample it at 11kHz, and randomly crop 6-second clips to generate $256\times 256$ spectrograms with log-frequency projection as input, then feed to U-Net.
To acquire effective visual guidance of sound source, we use clip-level audio tags for classification and perform audiovisual alignment within the contained 11 instruments. Then the visual representation of sound source is generated to guide source separation.

To precisely evaluate the separation performance, we adopt three metrics of Signal-to-Distortion Ratio (SDR), Signal-to-Interference Ratio (SIR) and Signal-to-Artifact Ratio (SAR), where higher is better for all \cite{sop,co-separation}.
Table~\ref{tab: music} shows separation results under different training conditions, where Single-Source means training with only solo videos, while Multi-Source refers to training with both solo and duet videos. We compare three different learning settings, the first is to directly use weights output by Grad-CAM as prediction mask, and the latter two are using audio and visual representation as guidance. The separation performance with Grad-CAM output weight is relatively poor, because it is of very low resolution, far from enough precise for sound separation. As for audio representations, since they are disentangled from mixed spectrogram by weighted pooling (Eq. \ref{con: pooling}), it is only slightly better than Grad-CAM but still not enough to represent a specific instrument. But when using visual representation as guidance, our model achieves comparable results on all three metrics. It demonstrates that our sound localization results contribute to effective visual representation of specific sound sources. Note that our model is trained with fewer audiovisual pairs compared to other methods, and \cite{co-separation} adopts an additional detector to extract sound source but which is not necessary for our model. To further validate the efficacy of our approach in multi-source scenes, our model is also trained with duet videos. The results reveal that our model can capture useful information in complex scenes to establish cross-modal association.
\begin{table}[]
    \centering
    \caption{Sound source separation results on MUSIC dataset. We report performance when training only on single-source (solo) videos and multi-source (solo+duet) videos as \cite{co-separation}. Note that SAR only captures absence of artifacts, and can be high even if separation of poor quality.}
    \begin{spacing}{1.00}
    \begin{tabular}{c|c c c|c c c}
    \hline
    \multirow{2}*{Methods} & \multicolumn{3}{c|}{Single-Source} & \multicolumn{3}{|c}{Multi-Source} \\
    \cline{2-7}
    ~ & SDR & SIR & SAR & SDR & SIR & SAR \\
    \hline
    NMF-MFCC\cite{nmf} & 0.92 & 5.68 & 5.84 & 0.92 & 5.68 & 5.84 \\
    AV-Mix-Sep\cite{MIML} & 3.16 & 6.74 & 8.89 & 3.23 & 7.01 & 9.14 \\
    Sound-of-Pixels\cite{sop} & 7.30 & 11.90 & 11.90 & 6.05 & 9.81 & 12.40 \\
    Co-Separation\cite{co-separation} & 7.38 & 13.70 & 10.80 & 7.64 & 13.80 & 11.30 \\
    CAVL\cite{curriculum} & 6.59 & 10.10 & 12.56 & 6.78 & 10.62 & 12.19 \\
    Ours Grad-CAM & -2.78 & -0.01 & 7.79 & -2.49 & 0.08 & 8.34 \\
    Ours Audio & -1.16 & 0.33 & 11.11 & -0.97 & 0.43 & 11.02 \\
    Ours Visual & 6.53 & 12.15 & 11.31 & 6.57 & 11.90 & 10.78 \\ 
    \hline
    \end{tabular}
    \end{spacing}
    \label{tab: music}
\end{table}

\section{Conclusions}

In this work, we present an audiovisual learning framework which automatically disentangles audio and visual representations of different categories from complex scenes, and performs feature alignment in a coarse-to-fine manner. We further propose a novel evaluation pipeline for multi-source sound localization to demonstrate the superiority of our model. And our model shows promising performance on sound localization in complex scenes with multiple sound sources, as well as on sound source separation.

In future, to better distinguish different sounds and objects, we would like to introduce more categories into classification task. In this way, we are able to establish more precise sound-object association.


\section{Acknowledgement}
The paper is supported in part by the following grants: China Major Project for New Generation of AI Grant (No.2018AAA0100400), National Natural Science Foundation of China (No. 61971277, No. 61901265). 

\clearpage
%
%
\bibliographystyle{splncs04}
\bibliography{egbib}
\end{document}


\pagestyle{headings}
\mainmatter
\def\ECCVSubNumber{3429}  
\newcommand{\etal}{et al. }
\newcommand{\ie}{i.e.}
\newcommand{\eg}{e.g.}
\newcommand{\etc}{etc.}
\title{Appendix} 

\section{Generating Pseudo Labels for Unlabelled Videos}

\label{sec: label}
When training on unlabelled videos, such as SonudNet-Flickr and AVE dataset, we need to generate pseudo labels as classification supervision.

First, we use CRNN pretrained on AudioSet and ResNet-18 pretrained on ImageNet to predict classification probabilities on audio and visual message.
Next, to reduce noise and assist coarse-grained audiovisual correspondence, we need to organize several general categories as target. Considering AudioSet is annotated with hierarchical ontology, containing four levels of labels from coarse to fine, we choose the first-level labels of 7 classes (human sounds, music, animal, sounds of things, natural sounds, source-ambiguous sounds, and environment) as final classification target. Then we aggregate the predictions from pretrained models into these 7 categories. For audio modality, we directly use the ontology in AudioSet to generate supervision. While for visual modality, we take similarity of word embeddings and conditional probabilities between labels in ImageNet and AudioSet into consideration to aggregate 1000 classification predictions into 7 as pseudo labels.

\section{Experiments on AVE Dataset}

\subsection{AVE Dataset}

AVE dataset contains 4143 10-second video clips covering 28 event categories. This dataset is proper for cross-modality localization since the videos are temporally labelled with audiovisual event boundaries. But annotations are only used for evaluation. In training phase, we  feed audiovisual pairs into our model to learn cross-modal alignment in an unsupervised manner. The videos are divided into 3339 for training, 402 for validation and 402 for test. Note that events in testing videos all span less than 10 seconds.

\subsection{Cross-Modality Localization}
In this task, given a temporal segment of one modality, we aim to accurately localize the temporal position of the synchronized content in the other modality. There are two subtasks, visual localization from audio segments and vice versa, namely A2V and V2A.
We adopt AVE dataset without labels for training, and only use short-event videos for evaluation.

Concretely, we employ sliding windows to predict the temporal position. Take visual localization from audio (A2V) as an example:
\begin{align}
    t^* = \arg\min_t\sum_{s=1}^l f(V_{s+t-1},\hat{A_s}),
\end{align}
where $f$ measures the correspondence score between audio and visual context, $\hat{A}$ represents query $l$-second audio segment, $t^*$ is the predicted start time when audio and vision synchronize. Strict evaluation metric is adopted on two subtasks. In Tabel~\ref{tab: ave}, we show our model's results in two different settings, one is only using classification and video-level audiovisual correspondence, the other is to further perform fine-grained alignment. Since it is more challenging to disentangle different events in mixed audio than in video frames, previous methods are poor on V2A. While our method performs much better at capturing temporal information in audio, and outperforms others over a large margin on V2A. Comparing results of our method with different settings, our fine-grained alignment in the second stage further improves performance, but still not the best on A2V. That is because the major target of this task is to distinguish temporal boundaries of audiovisual events, there are few events overlapping at the same time, which restricts the efficacy of our fine-grained alignment.
\setlength{\tabcolsep}{3mm}{
\begin{table}[]
    \centering
    \caption{Cross-madality localization accuracy with A2V and V2A subtasks.}
    \begin{spacing}{1.00}
    \begin{tabular}{c c c c c}
    \hline
    Models & DCCA & AVDLN  & Ours & Ours w/align \\
    \hline
    A2V & 34.8 & 44.8 & 41.5 & 43.8\\
    \hline
    V2A & 34.1 & 35.6 & 43.8 & 44.3\\
    \hline
    \end{tabular}
    \end{spacing}
    \label{tab: ave}
\end{table}}

\begin{figure}[]
    \centering
    \subfigure[From background noise to musical instruments.]{    \includegraphics[width=0.105\linewidth]{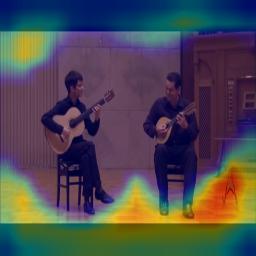}\hspace{-1mm}
    \includegraphics[width=0.105\linewidth]{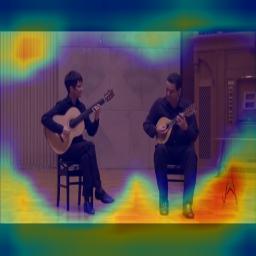}\hspace{-1mm}
    \includegraphics[width=0.105\linewidth]{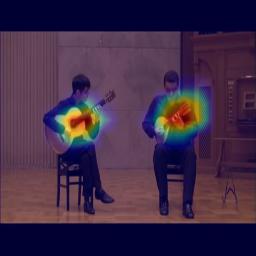}\hspace{-1mm}
    \includegraphics[width=0.105\linewidth]{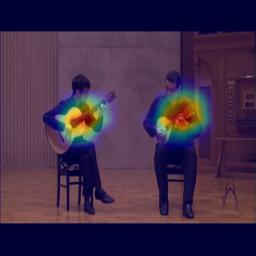}\hspace{-1mm}
    \includegraphics[width=0.105\linewidth]{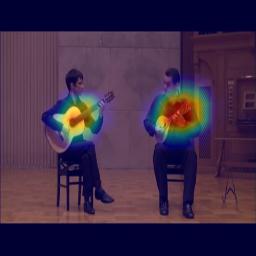}\hspace{-1mm}
    \includegraphics[width=0.105\linewidth]{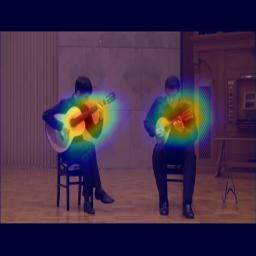}\hspace{-1mm}
    \includegraphics[width=0.105\linewidth]{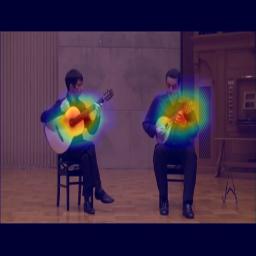}\hspace{-1mm}
    \includegraphics[width=0.105\linewidth]{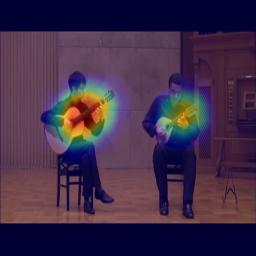}\hspace{-1mm}
    \includegraphics[width=0.105\linewidth]{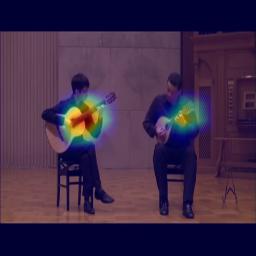}\hspace{-1mm}
    \label{fig: noise-music}}
    \subfigure[Duet of accordion and guitar.]{
    \includegraphics[width=0.105\linewidth]{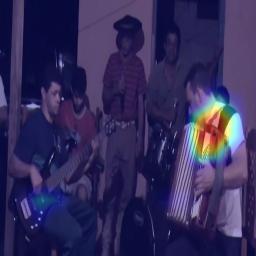}\hspace{-1mm}
    \includegraphics[width=0.105\linewidth]{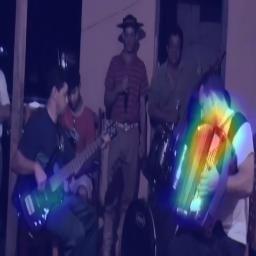}\hspace{-1mm}
    \includegraphics[width=0.105\linewidth]{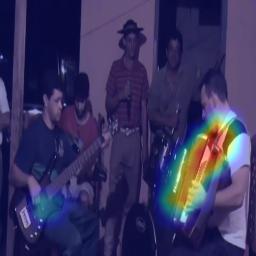}\hspace{-1mm}
    \includegraphics[width=0.105\linewidth]{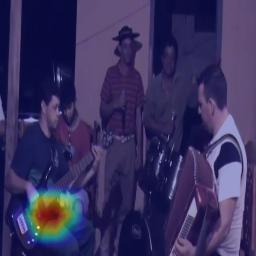}\hspace{-1mm}
    \includegraphics[width=0.105\linewidth]{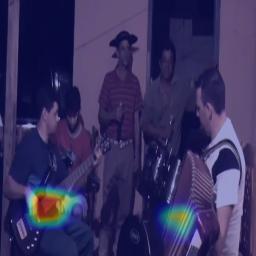}\hspace{-1mm}
    \includegraphics[width=0.105\linewidth]{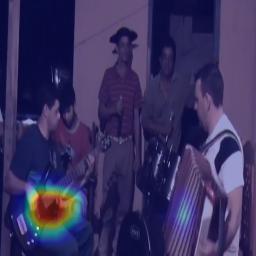}\hspace{-1mm}
    \includegraphics[width=0.105\linewidth]{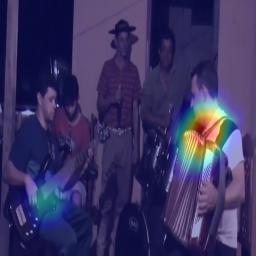}\hspace{-1mm}
    \includegraphics[width=0.105\linewidth]{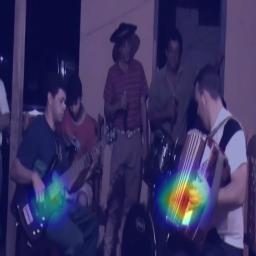}\hspace{-1mm}
    \includegraphics[width=0.105\linewidth]{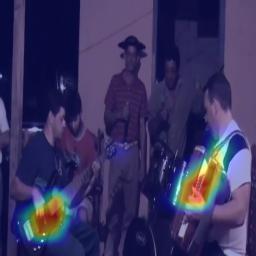}\hspace{-1mm}
    \label{fig: duet}}
    \subfigure[Dogs barking interspersed with sound of toy car.]{
    \includegraphics[width=0.105\linewidth]{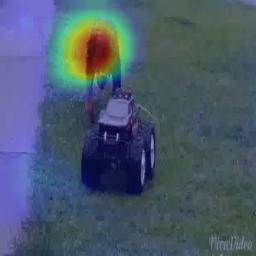}\hspace{-1mm}
    \includegraphics[width=0.105\linewidth]{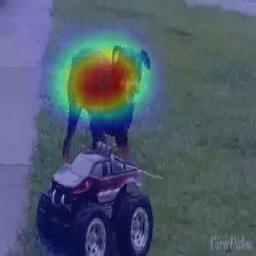}\hspace{-1mm}
    \includegraphics[width=0.105\linewidth]{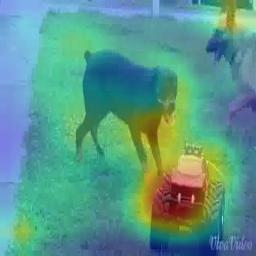}\hspace{-1mm}
    \includegraphics[width=0.105\linewidth]{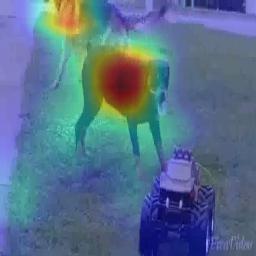}\hspace{-1mm}
    \includegraphics[width=0.105\linewidth]{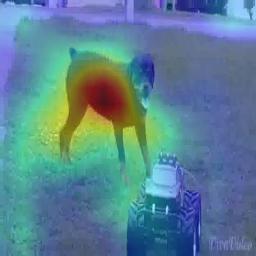}\hspace{-1mm}
    \includegraphics[width=0.105\linewidth]{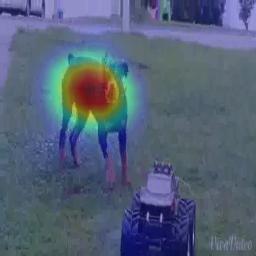}\hspace{-1mm}
    \includegraphics[width=0.105\linewidth]{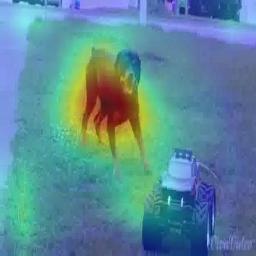}\hspace{-1mm}
    \includegraphics[width=0.105\linewidth]{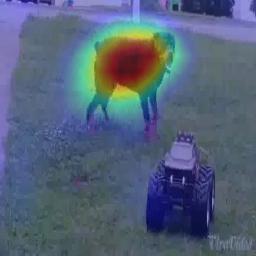}\hspace{-1mm}
    \includegraphics[width=0.105\linewidth]{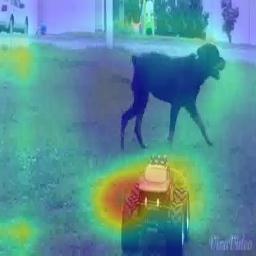}\hspace{-1mm}
    \label{fig: dog-car}}
    \caption{We visualize the changes of localization maps in videos over time. The frames shown are extracted at 1 fps, the heatmaps show localization responses to corresponding 1-second audio clip. When only with noise, our model mainly focuses on background regions as the first two frames in Fig.~\ref{fig: noise-music}. When there are sounds produced by specific objects, our model can accurately capture the sound makers, \eg, our model can distinguish sounds of guitar and accordion in Fig.~\ref{fig: duet}, dog barking and toy-car sound in Fig.~\ref{fig: dog-car}.}
    \label{fig:ave}
\end{figure}
We also visualize sound localizetion results on several videos. Fig.~\ref{fig:ave} vividly shows the changes of sounds on time dimension, which further demonstrates model's capacity of spatio-temporally determining which specific object is making sound.

\section{Comparison with CAM}

In this section, we compare the localization results between our model and CAM method based on classification.
Specifically, our two-stage framework achieves coarse-grained audiovisual correspondence in the category-level at the first stage, and fine-grained sound-object alignment at the second stage. 
To validate the efficacy of our fine-grained audiovisual alignment in the second stage, we compare our method with category-level CAM output.

\begin{figure}
    \centering
    \subfigure[background music]{
    \includegraphics[width=0.16\linewidth]{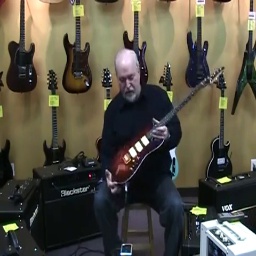}\hspace{-1mm}
    \includegraphics[width=0.16\linewidth]{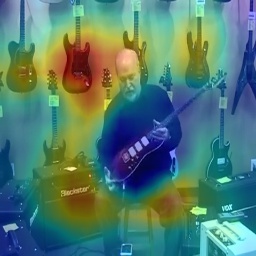}\hspace{-1mm}
    \includegraphics[width=0.16\linewidth]{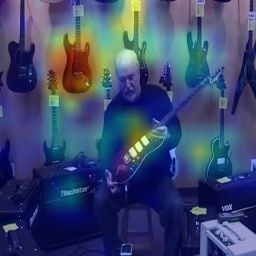}
    \label{fig: back}
    }\hspace{-2mm}
    \subfigure[playing guitar]{
    \includegraphics[width=0.16\linewidth]{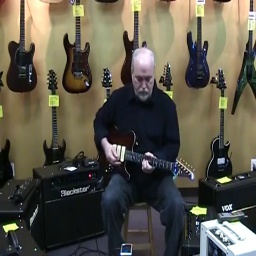}\hspace{-1mm}
    \includegraphics[width=0.16\linewidth]{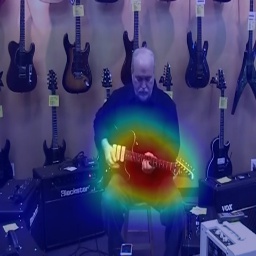}\hspace{-1mm}
    \includegraphics[width=0.16\linewidth]{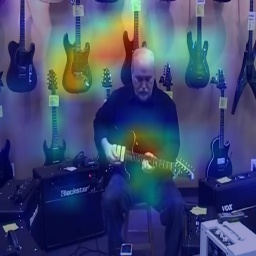}
    \label{fig: guitar}
    }
    \subfigure[rubbish truck]{
    \includegraphics[width=0.16\linewidth]{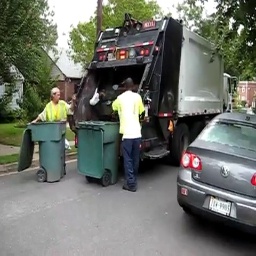}\hspace{-1mm}
    \includegraphics[width=0.16\linewidth]{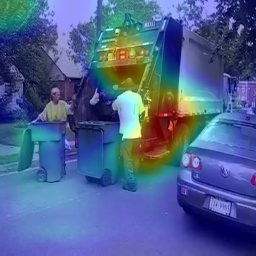}\hspace{-1mm}
    \includegraphics[width=0.16\linewidth]{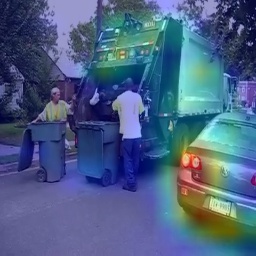}
    }\hspace{-2mm}
    \subfigure[aeroplane engine]{
    \includegraphics[width=0.16\linewidth]{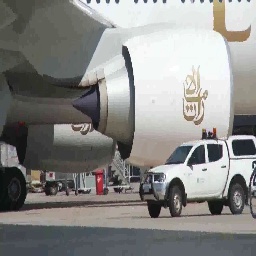}\hspace{-1mm}
    \includegraphics[width=0.16\linewidth]{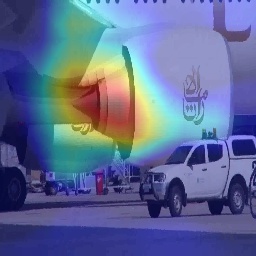}\hspace{-1mm}
    \includegraphics[width=0.16\linewidth]{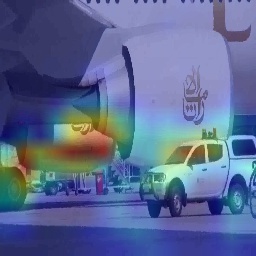}
    \label{fig: plcar}
    }
    \subfigure[playing violin]{
    \includegraphics[width=0.16\linewidth]{eccv2020kit/cmp/4_ori.jpg}\hspace{-1mm}
    \includegraphics[width=0.16\linewidth]{eccv2020kit/cmp/4_our.jpg}\hspace{-1mm}
    \includegraphics[width=0.16\linewidth]{eccv2020kit/cmp/4_cam.jpg}
    \label{fig: muins}
    }\hspace{-2mm}
    \subfigure[wood sawing]{
    \includegraphics[width=0.16\linewidth]{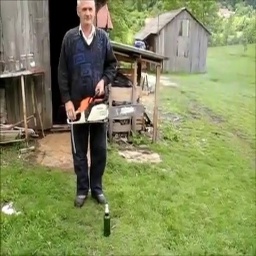}\hspace{-1mm}
    \includegraphics[width=0.16\linewidth]{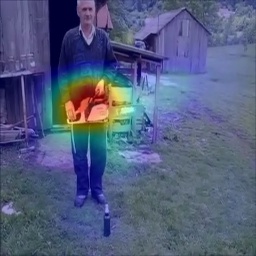}\hspace{-1mm}
    \includegraphics[width=0.16\linewidth]{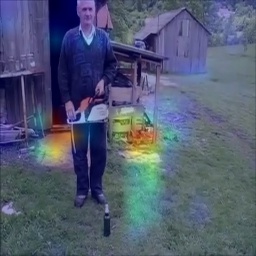}
    }
    \caption{We show some comparison between our model and CAM method. The images in each subfigure are listed as: original image, localization result of our model, result of CAM method. It is clear that CAM method cannot distinguish the objects belonging to the same category, \eg, violin and piano in Fig.~\ref{fig: muins}, but our model can precisely localize the object that makes sound in input audio.}
    \label{fig: comp}
\end{figure}

Concretely, we adopt the model trained on AVE dataset for comparison, where the classification targets are 7 general categories mentioned above (\ie, human sounds, music, animal, sounds of things, natural sounds, source-ambiguous sounds, and environment). 
Our model generates localization results following the procedure mentioned in the paper, while for CAM method, we adopt predicted probabilities on audio as prior, and employ CAM to generate class-specific localization maps on visual modality.
We visualize some comparison results in Fig.~\ref{fig: comp}. Generally, CAM method cannot distinguish the objects belonging the same category, \eg, aeroplane and car in Fig.~\ref{fig: plcar}, while our model can precisely localize the specific object making sound in input audio. It is because CAM method performs localization in the category-level, while our model further establishes video- and category-based sound-object association. Additionally, as shown in Fig.~\ref{fig: back} and Fig.~\ref{fig: guitar}, in the scene with multiple guitars, with background music sound, our model focuses on the silent guitars hanging on the wall, while with the sound of the man playing guitar, our method precisely localize the guitar held by the man. It is probably because the sound of playing guitar usually coexists with the visual pattern of the interaction between human hands with guitar, while the background music is usually with individually placed music instruments.

Further, we also quantitatively compare the localization results of these two methods on human annotated subset of SoundNet-Flickr dataset. For CAM method, we perform weighted summation on class activation maps over valid categories, where the weights are the normalized predicted probabilities on audio modality. Table \ref{tab: cmpflickr} shows the results, our two-stage learning framework outperforms CAM method over a large margin, which demonstrates the efficacy of fine-grained sound-object alignment in the second stage.

\setlength{\tabcolsep}{4mm}{
\begin{table}[]
    \centering
    \caption{Quantitative localization results on SoundNet-Flickr subset, cIoU and AUC are reported.}
    \begin{spacing}{1.00}
    \begin{tabular}{c | c c}
    \hline
    Methods & cIoU@0.5 & AUC  \\
    \hline
    Random & 7.2 & 30.7 \\
    Attention & 43.6 & 44.9 \\
    DMC AudioSet & 41.6 & 45.2 \\
    CAVL AudioSet & 50.0 & 49.2 \\
    Ours Stage-one & 44.2 & 48.1 \\
    Ours Stage-two & \textbf{52.2} & \textbf{49.6} \\
    \hline
    \end{tabular}
    \end{spacing}
    \label{tab: cmpflickr}
\end{table}}

\section{Additional Results}

\begin{figure}[t]
    \subfigure[speech with gunfire]{
    \includegraphics[width=0.105\linewidth]{eccv2020kit/flickr/3.jpg}\hspace{-1mm}
    \includegraphics[width=0.105\linewidth]{eccv2020kit/flickr/113_0.jpg}\hspace{-1mm}
    \includegraphics[width=0.105\linewidth]{eccv2020kit/flickr/113_3.jpg}
    \label{fig: human-gun}}
    \hspace{-4mm}
    \subfigure[cheering with engine]{
    \includegraphics[width=0.105\linewidth]{eccv2020kit/flickr/4.jpg}\hspace{-1mm}
    \includegraphics[width=0.105\linewidth]{eccv2020kit/flickr/203_0.jpg}\hspace{-1mm}
    \includegraphics[width=0.105\linewidth]{eccv2020kit/flickr/203_3.jpg}
    }
    \hspace{-4mm}
    \subfigure[music with inside noise]{
    \includegraphics[width=0.105\linewidth]{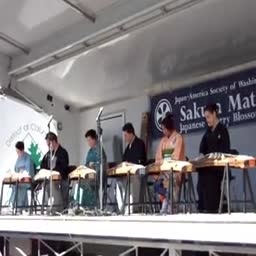}\hspace{-1mm}
    \includegraphics[width=0.105\linewidth]{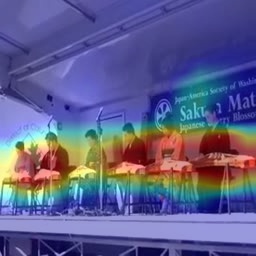}\hspace{-1mm}
    \includegraphics[width=0.105\linewidth]{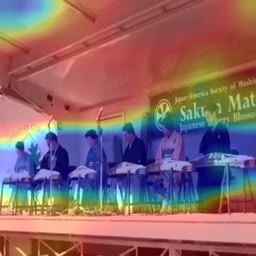}
    }
    \vspace{-3mm}\\
    \subfigure[shouting with water]{
    \includegraphics[width=0.105\linewidth]{eccv2020kit/flickr/0.jpg}\hspace{-1mm}
    \includegraphics[width=0.105\linewidth]{eccv2020kit/flickr/22_0.jpg}\hspace{-1mm}
    \includegraphics[width=0.105\linewidth]{eccv2020kit/flickr/22_3.jpg}
    }
    \hspace{-4mm}
    \subfigure[human with wind]{
    \includegraphics[width=0.105\linewidth]{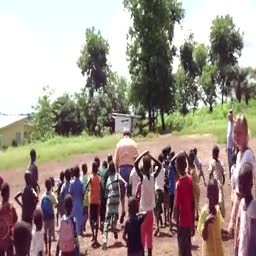}\hspace{-1mm}
    \includegraphics[width=0.105\linewidth]{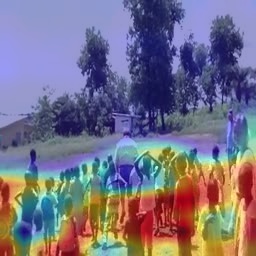}\hspace{-1mm}
    \includegraphics[width=0.105\linewidth]{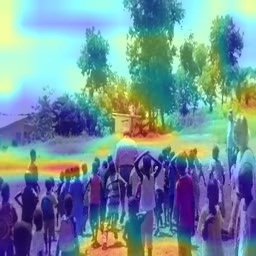}
    }
    \hspace{-4mm}
    \subfigure[sports with stadium]{
    \includegraphics[width=0.105\linewidth]{eccv2020kit/flickr/2.jpg}\hspace{-1mm}
    \includegraphics[width=0.105\linewidth]{eccv2020kit/flickr/93_2.jpg}\hspace{-1mm}
    \includegraphics[width=0.105\linewidth]{eccv2020kit/flickr/93_4.jpg}
    }
    \vspace{-3mm}\\
    \subfigure[sports with cheering]{
    \includegraphics[width=0.105\linewidth]{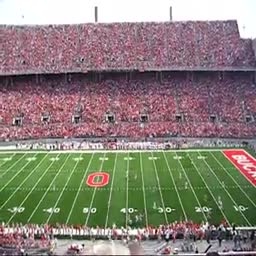}\hspace{-1mm}
    \includegraphics[width=0.105\linewidth]{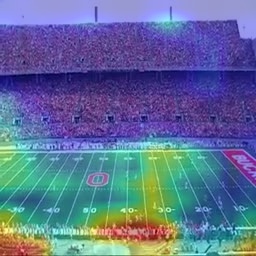}\hspace{-1mm}
    \includegraphics[width=0.105\linewidth]{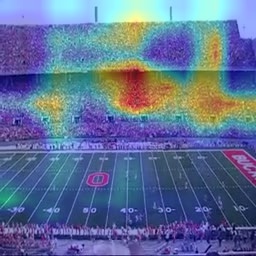}
    }
    \hspace{-4mm}
    \subfigure[speech with motorcycle]{
    \includegraphics[width=0.105\linewidth]{eccv2020kit/flickr/30.jpg}\hspace{-1mm}
    \includegraphics[width=0.105\linewidth]{eccv2020kit/flickr/30_0.jpg}\hspace{-1mm}
    \includegraphics[width=0.105\linewidth]{eccv2020kit/flickr/30_3.jpg}
    }
    \hspace{-4mm}
    \subfigure[engine with wind]{
    \includegraphics[width=0.105\linewidth]{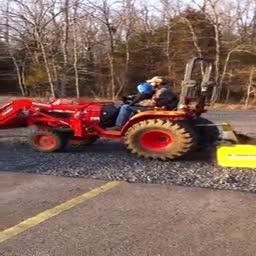}\hspace{-1mm}
    \includegraphics[width=0.105\linewidth]{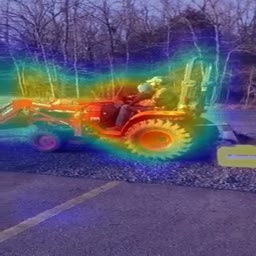}\hspace{-1mm}
    \includegraphics[width=0.105\linewidth]{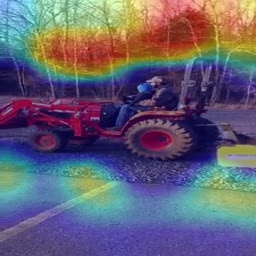}
    }
    \vspace{-3mm}\\
    \subfigure[yelling with impact]{
    \includegraphics[width=0.105\linewidth]{eccv2020kit/flickr/227.jpg}\hspace{-1mm}
    \includegraphics[width=0.105\linewidth]{eccv2020kit/flickr/227_0.jpg}\hspace{-1mm}
    \includegraphics[width=0.105\linewidth]{eccv2020kit/flickr/227_1.jpg}
    \label{fig: human-impact}}
    \hspace{-4mm}
    \subfigure[human with dog]{
    \includegraphics[width=0.105\linewidth]{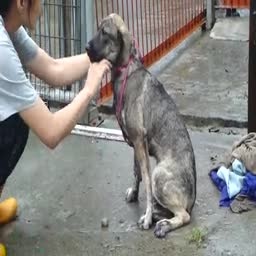}\hspace{-1mm}
    \includegraphics[width=0.105\linewidth]{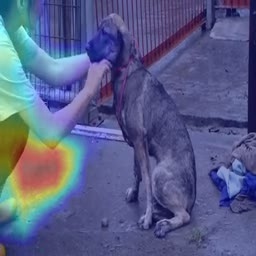}\hspace{-1mm}
    \includegraphics[width=0.105\linewidth]{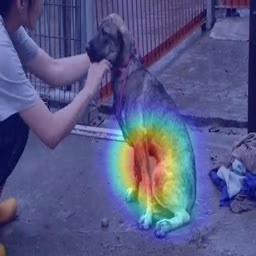}
    }
    \hspace{-4mm}
    \subfigure[talking with water]{
    \includegraphics[width=0.105\linewidth]{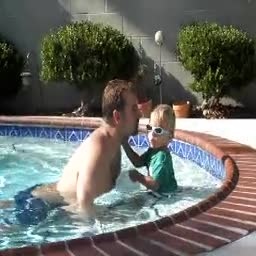}\hspace{-1mm}
    \includegraphics[width=0.105\linewidth]{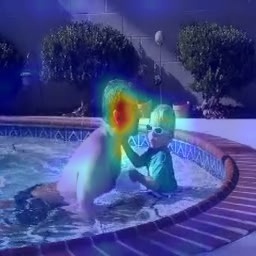}\hspace{-1mm}
    \includegraphics[width=0.105\linewidth]{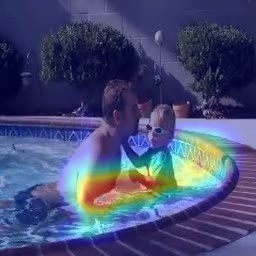}
    }
    \vspace{-3mm}\\
    \subfigure[screaming with stadium]{
    \includegraphics[width=0.105\linewidth]{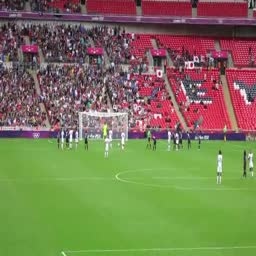}\hspace{-1mm}
    \includegraphics[width=0.105\linewidth]{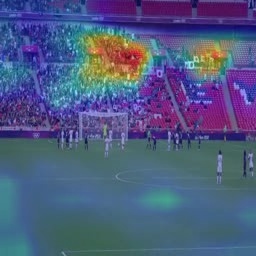}\hspace{-1mm}
    \includegraphics[width=0.105\linewidth]{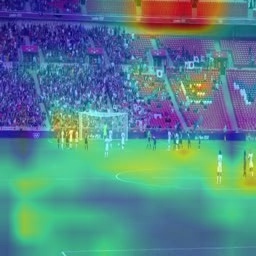}
    }
    \hspace{-4mm}
    \subfigure[yelling with wind]{
    \includegraphics[width=0.105\linewidth]{eccv2020kit/flickr/211.jpg}\hspace{-1mm}
    \includegraphics[width=0.105\linewidth]{eccv2020kit/flickr/211_0.jpg}\hspace{-1mm}
    \includegraphics[width=0.105\linewidth]{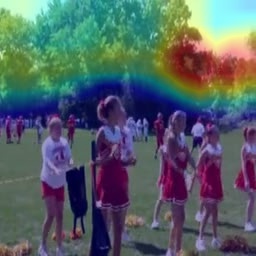}
    }
    \hspace{-4mm}
    \subfigure[speech with classroom]{
    \includegraphics[width=0.105\linewidth]{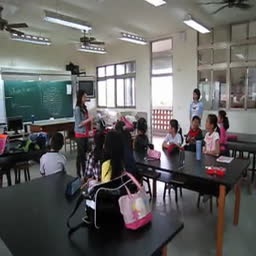}\hspace{-1mm}
    \includegraphics[width=0.105\linewidth]{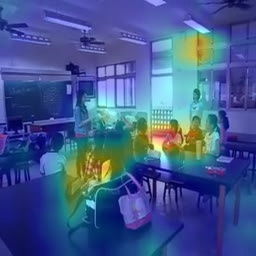}\hspace{-1mm}
    \includegraphics[width=0.105\linewidth]{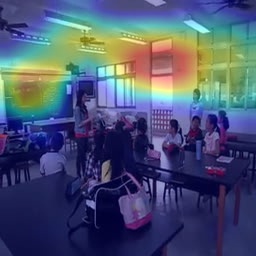}
    }
    \caption{We visualize the localization maps corresponding to different elements contained in the mixed sounds of two sources. The results qualitatively demonstrate our model's performance in multi-source sound localization.}
    \label{fig:flickr}
\end{figure}

In this section, we present more examples of our localization results in multi-source scenarios. Fig. \ref{fig:flickr} shows the result in two-source scenes, and the results generally demonstrate our model's capacity of distinguishing different sound sources.

\begin{figure}
    \centering
    \subfigure[shouting, engine and water]{
    \includegraphics[width=0.12\linewidth]{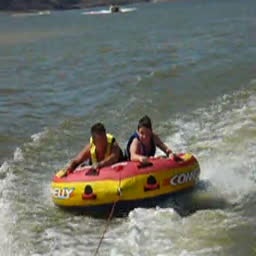}\hspace{-1mm}
    \includegraphics[width=0.12\linewidth]{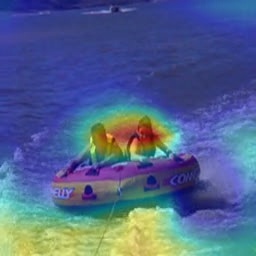}\hspace{-1mm}
    \includegraphics[width=0.12\linewidth]{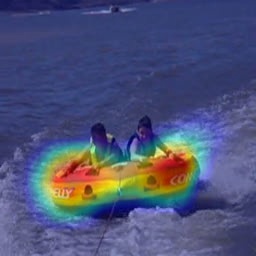}\hspace{-1mm}
    \includegraphics[width=0.12\linewidth]{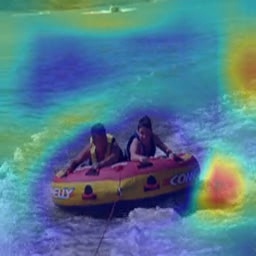}
    \label{fig: sew}}
    \hspace{-3mm}
    \subfigure[speaking, gunfire and wind]{
    \includegraphics[width=0.12\linewidth]{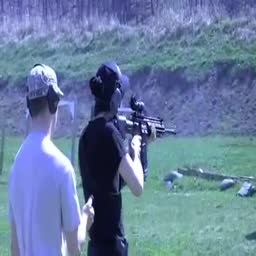}\hspace{-1mm}
    \includegraphics[width=0.12\linewidth]{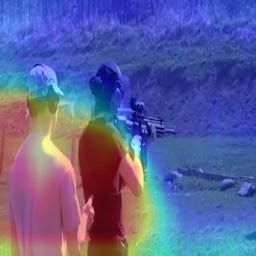}\hspace{-1mm}
    \includegraphics[width=0.12\linewidth]{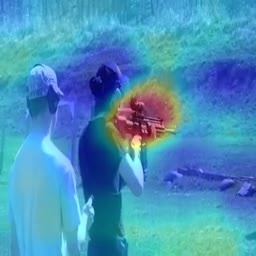}\hspace{-1mm}
    \includegraphics[width=0.12\linewidth]{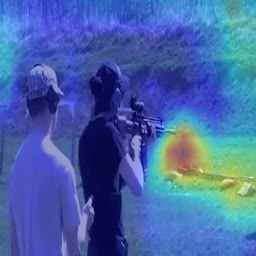}
    \label{fig:cam}}
    \caption{We show the localization results of three-source scenes, and each localization map corresponds to one potential sound source.}
    \label{fig:3-source}
\end{figure}

We also show some localization results under three-source scenes in Fig. \ref{fig:3-source}. In Fig. \ref{fig: sew}, it is interesting that the boat is being towed by something off-screen, and the engine sound actually comes from the unseen object, while our model associates them as a sound-object pair. This is probably because the visual pattern of boats usually coexist with engine sound, and these two are of the same category, eventually they become highly correlated. 

\begin{figure}
    \centering
    \subfigure[sound of baby as query and top-5 retrieved images]{
    \includegraphics[width=0.16\linewidth]{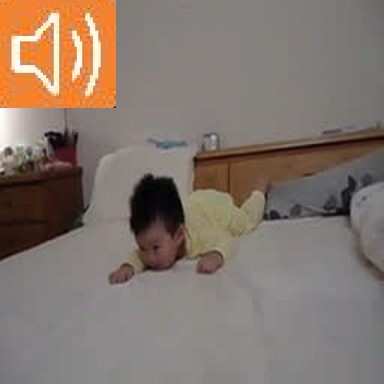}\hspace{1mm}
    \includegraphics[width=0.16\linewidth]{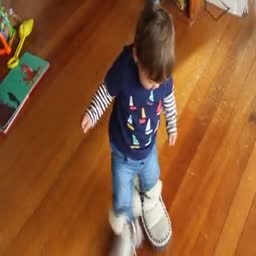}\hspace{-1mm}
    \includegraphics[width=0.16\linewidth]{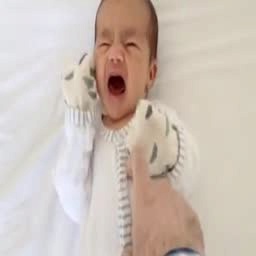}\hspace{-1mm}
    \includegraphics[width=0.16\linewidth]{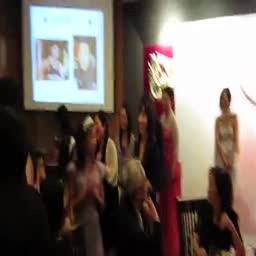}\hspace{-1mm}
    \includegraphics[width=0.16\linewidth]{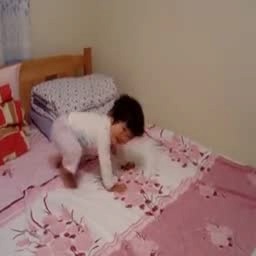}\hspace{-1mm}
    \includegraphics[width=0.16\linewidth]{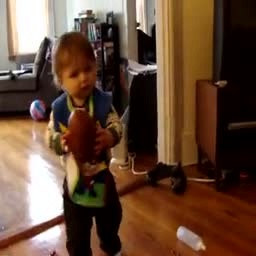}
    }\\
    \subfigure[sound of helicopter engine as query and top-5 retrieved images]{
    \includegraphics[width=0.16\linewidth]{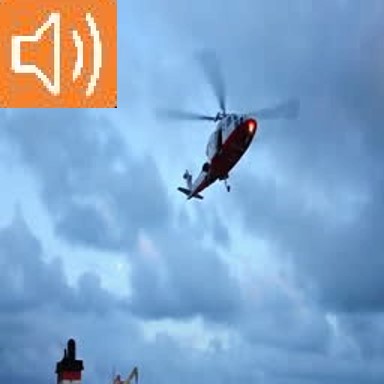}\hspace{1mm}
    \includegraphics[width=0.16\linewidth]{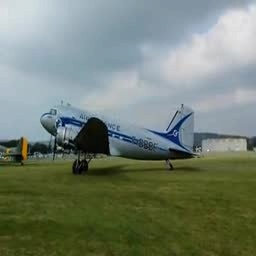}\hspace{-1mm}
    \includegraphics[width=0.16\linewidth]{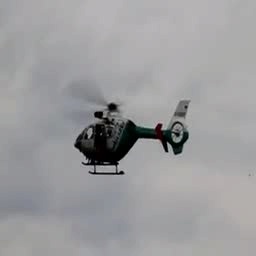}\hspace{-1mm}
    \includegraphics[width=0.16\linewidth]{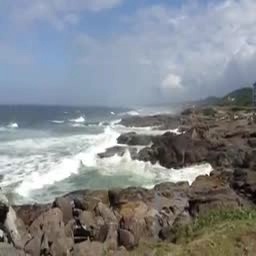}\hspace{-1mm}
    \includegraphics[width=0.16\linewidth]{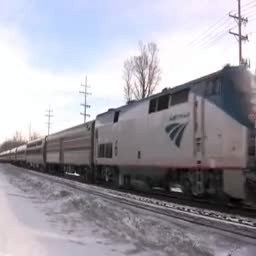}\hspace{-1mm}
    \includegraphics[width=0.16\linewidth]{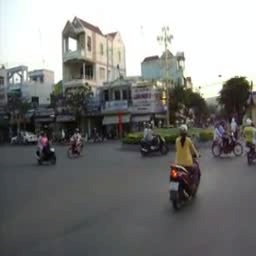}
    }\\
    \subfigure[image of crowd people as query and top-5 retrieved audio]{
    \includegraphics[width=0.16\linewidth]{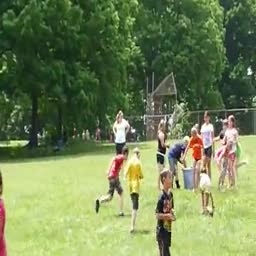}\hspace{1mm}
    \includegraphics[width=0.16\linewidth]{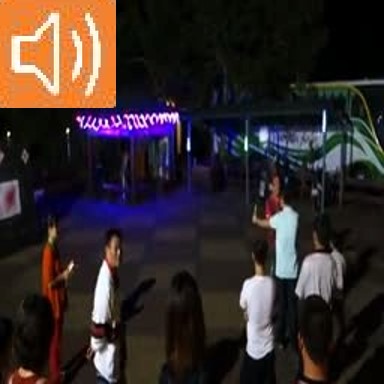}\hspace{-1mm}
    \includegraphics[width=0.16\linewidth]{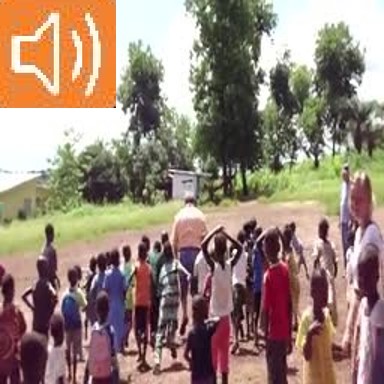}\hspace{-1mm}
    \includegraphics[width=0.16\linewidth]{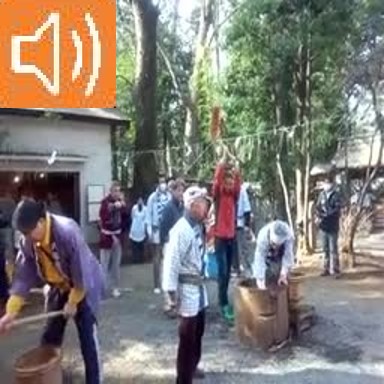}\hspace{-1mm}
    \includegraphics[width=0.16\linewidth]{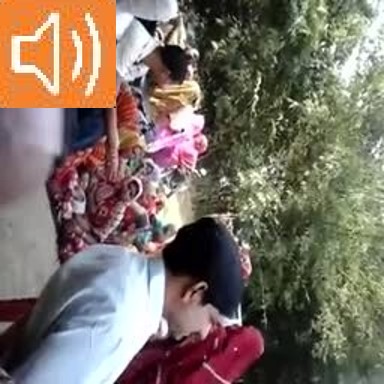}\hspace{-1mm}
    \includegraphics[width=0.16\linewidth]{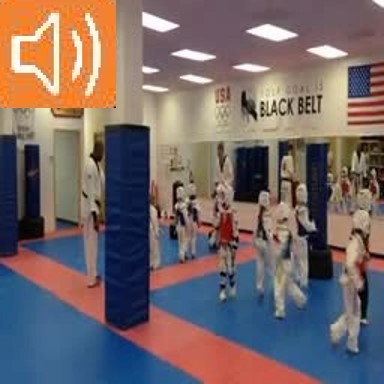}
    }
    \caption{Cross-modal retrieval results, with an image/sound as query and retrieve top-5 most similar audio/images.}
    \label{fig:retrieve}
\end{figure}
We present cross-modal retrieval results based on the aligned audiovisual features in Fig. \ref{fig:retrieve}. Concretely, we use an image or a clip of audio as query, and treat other audio or images in the dataset as gallery. We calculate the distance between query and gallery features, and take the top-5 nearest examples shown in Fig. \ref{fig:retrieve}.
\clearpage
%
%
\bibliographystyle{splncs04}
\bibliography{egbib}